\pdfoutput=1

\documentclass[11pt]{article}

\usepackage{emnlp2021}

\usepackage{times}
\usepackage{latexsym}

\usepackage[T1]{fontenc}

\usepackage[utf8]{inputenc}

\usepackage{microtype}
\usepackage{multirow}

\usepackage{graphicx}
\usepackage{amsmath}
\usepackage{hyperref}
\usepackage{stfloats}
\usepackage{float}
\usepackage{placeins}
\restylefloat{table}
\usepackage[normalem]{ulem}
\useunder{\uline}{\ul}{}
\usepackage{amssymb}
\usepackage{pifont}
\usepackage[colorinlistoftodos]{todonotes}
\usepackage{booktabs}

\title{Exploring a Unified Sequence-To-Sequence Transformer for\\
Medical Product Safety Monitoring in Social Media}

\author{
  Shivam Raval$^{1}$ \thanks{\hspace{0.1cm} equal contribution}\quad 
  Hooman Sedghamiz$^1$ \footnotemark[1] \quad
  \textbf{Enrico Santus}$^1$ \quad
  \textbf{Tuka Alhanai}$^2$ \quad \\
  \textbf{Mohammad Ghassemi}$^3$ \quad \textbf{Emmanuele Chersoni}$^4$ \\
  $^1$ DSIG - Bayer Pharmaceuticals, New Jersey, USA \\
  $^2$ New York University, Abu Dhabi, UAE \\
  $^3$ Michigan State University, Michigan, USA \\
  $^4$ The Hong Kong Polytechnic University, Hong Kong \\
  \texttt{sbraval@asu.edu}, \texttt{\{hooman.sedghamiz,enrico.santus\}@bayer.com}\\
  \texttt{tuka.alhanai@nyu.edu}, \texttt{ghassem3@msu.edu} \\ 
  \texttt{emmanuele.chersoni@polyu.edu.hk}
}

\begin{document}
\maketitle
\begin{abstract}
Adverse Events (AE) are harmful events resulting from the use of medical products. Although social media may be crucial for early AE detection, the sheer scale of this data makes it logistically intractable to analyze using human agents, with NLP representing the only low-cost and scalable alternative.

In this paper, we frame AE Detection and Extraction as a sequence-to-sequence problem using the T5 model architecture and achieve strong performance improvements over competitive baselines on several English benchmarks (F1 = 0.71, 12.7\% relative improvement for AE Detection; Strict F1 = 0.713, 12.4\% relative improvement for AE Extraction). Motivated by the strong commonalities between AE-related tasks, the class imbalance in AE benchmarks and the linguistic and structural variety typical of social media posts, we propose a new strategy for multi-task training that accounts, at the same time, for task and dataset characteristics. Our multi-task approach increases model robustness, leading to further performance gains. Finally, our framework shows some language transfer capabilities, obtaining higher performance than Multilingual BERT in zero-shot learning on French data.   
\end{abstract}

\section{Introduction}
Before market release, drugs are regularly tested for safety and effectiveness in clinical trials. However, since no clinical trial is large enough to find all potential Adverse Events (AEs) on a wide and diverse range of population, Pharmacovigilance continuously monitors the market to timely intervene, in case unexpected AEs are discovered. 

According to multiple sources \citep{sen2016consumer,alatawi2017empirical}, AEs are systematically under-reported in official channels. A growing number of patients, though, talk about them on social platforms like Twitter and health forums, sharing medical conditions, treatment reviews, side effect descriptions and so on. These outlets contain crucial information for Pharmacovigilance, but the sheer scale of this data -- velocity, volume, variety -- makes manual exploration prohibitively expensive. For this reason, Natural Language Processing (NLP) technologies represent the only low-cost and scalable alternative.

In recent years, the research community approached this problem by promoting thematic workshops and shared tasks, such as the Social Media Mining For Health Applications (SMM4H) \cite{weissenbacher2019overview,klein2020overview}, as well as by creating resources, such as CADEC \cite{karimi2015cadec}. Despite these efforts, the automatic detection of AEs from social outlets has still to face major challenges: i) posts containing AEs are rare compared to other posts (i.e. \textbf{rare signal and imbalanced data}); ii) text typologies largely differ across media (i.e. \textbf{text length and structure}); iii) \textbf{informal and figurative language} is dominant, often containing slang, idioms, sarcasm and metaphors; iv) datasets contain broad \textbf{differences in the annotations}, sometimes focusing only on the symptom mentions and others times including temporal, locative and intensity modifiers; v) annotated resources for model fine-tuning are only available for a small set of languages (i.e. \textbf{cross-lingualism}). 

Most of these challenges have led the research community to develop end-to-end solutions for each task, missing the benefit of performing multiple related tasks with the same model (i.e. transfer learning). In this paper, we aim to tackle the above-mentioned challenges all at once by framing the AE detection and AE extraction tasks as generative sequence-to-sequence (seq-to-seq) problems, to be addressed with a single architecture, namely T5 \citep{raffel2019exploring}.
In previous studies, the T5 architecture showed high flexibility in dealing with text from different domains and typologies, even in knowledge-intensive tasks \citep{petroni2020kilt}.  Furthermore, the T5 architecture is capable of incrementally learning new tasks with few or no labels \citep{xue2020mt5}. 

In the following paragraphs, not only we show that T5 outperforms strong baselines on multiple English benchmarks
(F1 = 0.71, 10.94\% relative improvement for AE detection; Strict F1 = 0.713, 12.46\% relative improvement for AE extraction), but, to fully unleash its potential and thereby address the above-mentioned challenges (i.e., small, varied and imbalanced data), we introduce a novel multi-task/data training framework that efficiently handles task complexity, data imbalance and textual differences, further improving over the state-of-the-art results. Assessed in multiple cross-textual and (zero-shot learning) cross-lingual AE detection and AE extraction settings, T5 shows robustness and improves over all the competitive baselines, despite being simpler -- in terms of number of layers and parameters -- than the competitors.



To summarize, our contributions are:
i) we use T5 for framing AE detection and extraction as a sequence-to-sequence problem, obtaining strong performance on multiple tasks and datasets; 
ii) we describe a new approach for balancing data across tasks and datasets in a multi-task setting, which leads to F1-score improvements on all benchmarks; 
iii) we test our model in a crosslingual transfer (English to French) scenario, showing that it outperforms Multilingual BERT in zero-shot learning.

\begin{figure*}
\begin{center}
   \includegraphics[width=0.67\textwidth, height=3.8cm]{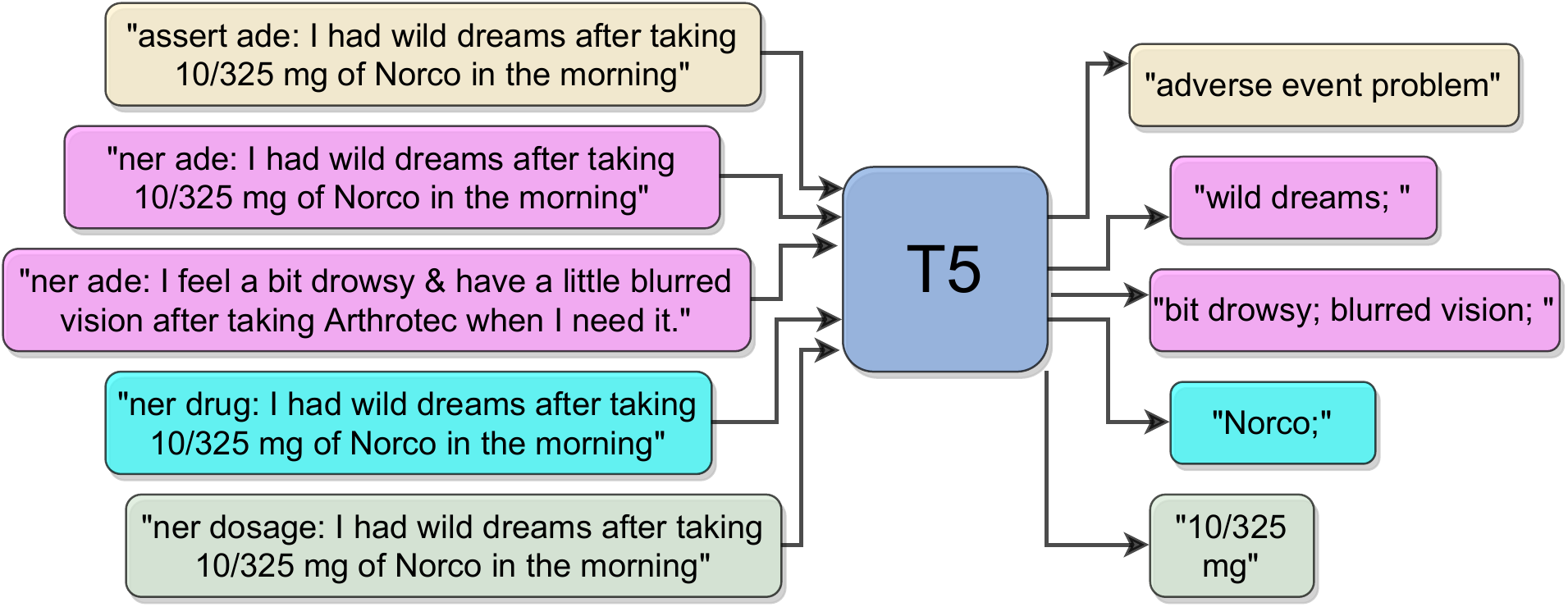}
\end{center}
   \caption{Diagram of our sequence-to-sequence framework, which is a fine-tuned T5 model for four prefix: ``assert ade:'' (yellow box) for detection task, ``ner ade:'' (pink box) for the task of extracting AE's , ``ner drug:'' (blue box) for extracting drug mentions and ``ner dosage:'' (green box) for extracting drug dosage information from the input.}
\label{fig:long1} 
\end{figure*}


\section{Related Work}
Early efforts in automated Pharmacovigilance have targeted Electronic Health Records (EHR) to detect evidence of AEs \cite{uzuner20112010, jagannatha2019overview}. However, not all AEs lead to clinical visitations: many users prefer to discuss their experiences with drugs on the Internet, and this fact led to a growing interest in the automatic detection of adverse events from social media platforms. 

Some of the early machine learning systems for AE detection from social media data used a combination of various classifiers along with word embeddings as features \cite{sarker2015portable,nikfarjam2015pharmacovigilance,daniulaityte2016bad,metke2016concept}. 

After the introduction of challenges such as Social Media Mining for Health Applications (SMM4H) \cite{weissenbacher2018overview, weissenbacher2019overview} and CADEC \cite{karimi2015cadec}, most works focused on neural networks \cite{sarker2018data, minard2018irisa}. With the development of attention mechanism \cite{vaswani2017attention}, Transformer-based language models such as BERT \cite{devlin2018bert} and its biomedical (e.g., BioBERT \citep{lee2020biobert}, ClinicalBERT \citep{alsentzer2019publicly} and PubMedBERT \citep{gu2020domain}) and non-biomedical variants (e.g., SpanBERT \citep{joshi2020spanbert}) obtained state-of-the-art performance in AE detection \citep{weissenbacher2019overview,klein2020overview,portelli-etal-2021-bert,beatrice2021span}.  

Models like BERT and its variants can be described as encoder-only: in order to carry out a specific task, a decoder has to be followed by task-specific trainable network, most often in the form of a linear layer. Recent developments in NLP led to the introduction of models such as T5 \cite{raffel2019exploring}, which is an encoder-decoder Transformer architecture. In a series of studies, T5 and its variants have shown performance gain on various datasets and applications \cite{raffel2019exploring,xue2020mt5}, despite being smaller in terms of parameters. The prefix training approach adopted by T5 allows users to fine-tune on various tasks concurrently, creating a single model that can incrementally learn while being capable of performing different tasks simultaneously. 
To our knowledge, the present contribution is the first to frame AE detection and extraction as generative problems. 

\section{Methods}
\subsection{The T5 Model}
We employ T5, a pre-trained encoder-decoder transformer proposed by~\citet{raffel2019exploring}. This model maps a vector sequence of $n$ input words represented by $\mathbf{X_{1:n}} = \mathbf{x_1,\cdots,x_n}$ to an output sequence of $\mathbf{Y_{1:m}} = \mathbf{y_1},\cdots,\mathbf{y_m}$ with an \textit{a-priori} unknown length of $m$, with a conditional probability defined as:

\begin{equation}
    p_{\theta_{model}}(\mathbf{Y}_{1:m}|\mathbf{X}_{1:n})
\end{equation}

The architecture of the model is very similar to the original Transformer proposed by~\citet{vaswani2017attention}. An input sequence is first passed to the encoder which consists of self-attention followed by feed-forward layers. The encoder maps the input to a sequence of embeddings that go through normalization and drop-out layers. The decoder attends to the output of the encoder using several attention layers. The self-attention layers, instead, employ masking to make the decoder only attend to the past tokens, in an auto-regressive manner:

\begin{equation}
    p_{\theta_{decoder}}(\mathbf{Y}_{1:m}) = \prod_{i=1}^{m}p_{\theta_{decoder}}(\mathbf{y_i}|\mathbf{Y}_{0:i-1})
\end{equation}
where $p_{\theta_{decoder}}(\mathbf{y_i}|\mathbf{Y}_{0:i-1})$ is the probability distribution of the next token $\mathbf{y}_i$.
Finally, the output of the decoder passes through a SoftMax layer over the vocabulary. 
\citet{raffel2019exploring} proposed to add a prefix in front of the input sequence to inform the model about which task to perform (e.g. summarization, question answering, classification etc.; see Figure~\ref{fig:long1}).
The model was trained on the Colossal Clean Crawled Corpus (C4), a massive corpus (about 750 GB) of web-extracted and cleaned text.

\subsubsection{Pre-Training and Pre-Finetuning}

\citet{raffel2019exploring} explored a wide range of architectures and pre-training objectives, finding that encoder-decoder models generally outperform decoder-only language models, and that a BERT-style denoising objective -- where the model is trained to recover masked words in the input -- works best.  Moreover, the best variant of their system made use of an objective that corrupts contiguous spans of tokens, similarly to the span corruption strategy introduced for the SpanBERT model \citep{joshi2020spanbert}.

The resulting model was then pre-finetuned on a variety of tasks from the following sources: the GLUE \citep{wang2018glue} and the SuperGLUE \citep{wang2019superglue} benchmarks for natural language understanding, the abstractive summarization data by \citet{hermann2015teaching} and \citet{nallapati2016abstractive}, the SQUAD question answering dataset \citep{rajpurkar2016squad} and the WMT translation benchmarks for translation from English to French, from English to German and from English to Romanian. The tasks were all treated as a single task in the sequence-to-sequence format, by concatenating all the datasets together and appending the task-specific prefixes to the instances. 

T5 comes in versions, \textit{small} (60 million parameters), \textit{base} (220 million parameters), \textit{large} (770 million parametrs), \textit{3B} (3 billion parameters) and \textit{11B} (11 billion parameters). In the paper we will use the term T5 to either refer to the architecture or to the T5-Base, as opposed to T5-Small, which will always be mentioned as such.



\subsection{Seq-to-Seq AE-related Tasks}

Given an input sequence of words $\mathbf{X_{1:n}} = \mathbf{x_1,\cdots,x_n}$ that potentially contains drug, dosage and AE mentions, we frame the AE detection (i.e. binary classification) and extraction (i.e. span detection) tasks as seq-to-seq problems, further fine-tuning T5 to generate $\mathbf{Y_{1:m}} = \mathbf{y_1},\cdots,\mathbf{y_m}$, where $Y$ is either the classification label or the text span with the AE. By selecting the prefixes (see Table \ref{table:tasks}), we train T5 on all these tasks (see Figure \ref{fig:long1}). 

\begin{table}[!ht]
\begin{small}
\centering
\resizebox{\columnwidth}{!}{%
\begin{tabular}{lll} 
\textbf{Prefix}   & \textbf{Task Definition}  & \textbf{Task Type}\\ 
\hline
\hline
\textit{assert ade} & Contains AE or not & CLS (binary)\\

\textit{ner ade} & Extract AE span & NER (span)\\

\textit{ner drug} & Extract drug span & NER (span)\\

\textit{ner dosage} & Extract drug dosage span & NER (span)\\

\end{tabular}
}
\end{small}
\caption{Prefix and task definition. AE assertion is binary classification (CLS), while the remaining tasks are Name Entity Recognition (NER).} 
\label{table:tasks}
\end{table}
For the AE detection, as it is a binary classification task, we have chosen the prefix ``assert ade:'' and the labels i) ``adverse event problem'' (i.e., positive) and ii) ``health ok'' (i.e., negative). Usually, to train Named Entity Recognition (NER) systems, the input data is transformed into standard Inside–Outside–Beginning (IOB) format and individual tokens are classified in one of the IOB tags. However, the T5 model can utilize the direct span as a generation target. If multiple spans can be extracted, they can be provided to the system separated by a semicolon or other special characters. 
For our experiments on language transfer, we simply apply the ``assert ade: '' to data in a different language (i.e., French). The model will automatically leverage the knowledge acquired during the \textit{pre-finetuning} in the machine translation task. 
Tasks and definitions are summarized in Table~\ref{table:tasks}.

\subsection{Multi-Task and Multi-Dataset Fine-Tuning}
\label{sec:training_strategies}
Generative models like T5 can be easily trained on multiple tasks. However, multi-task learning poses challenges as models may overfit or underfit, depending on the task difficulty, the label distribution and the variability across tasks and datasets \citep{arivazhagan2019massively}. In \citet{raffel2019exploring}, proportional mixing and temperature scaling training strategies were adopted to address the data balance across tasks. 
In this work, we extend these strategies to a \textbf{multi-dataset scenario}, in which tasks are trained on multiple datasets containing heterogeneous data. This scenario is typical in AE detection, where data comes from medical blogs, forums, tweets and other social media outlets, each of which carries specific writing styles as well as different textual structures and lengths. The annotation scheme may differ too across datasets, with some schemes focused on the symptoms only, while others including also the temporal, manner and intensity modifiers. 


We assume a multinomial probability distribution $\theta_t$ over the fine-tuning task $t$, given that the fine-tuning task $t$ itself is comprised of dataset(s) $d$. We define $M_d$ as the number of samples of dataset $d$ and $\rho_d$ the probability of drawing an example from $d$ during training.

In \emph{proportional mixing}, we intuitively sample in proportion to the dataset size. Therefore, the probability of drawing a sample from task $t$ is computed as $\theta_t = \frac{min(\gamma_t, N_t)}{\sum_{t} min(\gamma_t, N_t)}$, where $N_t$ corresponds to the number of samples available for task $t$ across all datasets, computed as $N_t=\sum_{d} M_d$. The probability of drawing from dataset $d$ is similarly estimated as $ \rho_d = \frac{min(\gamma_d, M_d)}{\sum_{t} min(\gamma_d, M_d)}$. For the sake of algorithm re-utilization, these parameters $\gamma_d$ and $\gamma_t$ are introduced because, even with proportional mixing, large datasets may still dominate the training. These parameters are meant to limit the impact of such large datasets and they have been set to $2^{14}$ as in the original paper \citep{raffel2019exploring}.

\emph{Temperature scaling} has also been shown to boost multi-task training performance \citep{raffel2019exploring,goodwin2020towards}. It was used for Multilingual BERT, to make sure that the model had sufficient training on low-resource languages \citep{devlin2018bert}. To implement scaling with a temperature $\mathcal{T}$, the mixing rate for each task and dataset is raised to the power of $1/\mathcal{T}$, and then the rates are re-normalized so that they sum to 1. Therefore, initially, the probabilities are computed with temperature scaling, respectively, as
$
\theta_t = \frac{\sqrt[\mathcal{T}]{\theta_t}}{\sum_{t} \sqrt[\mathcal{T}]{\theta_t} }    
$ (for the probability of drawing from task $t$) and as
$
\rho_d = \frac{\sqrt[\mathcal{T}]{\rho_d}}{\sum_{d} \sqrt[\mathcal{T}]{\rho_d} }
$ (for the probability of drawing from dataset $d$). We set $T$ as 2 as it is the best reported value for temperature scaling strategy demonstrated in \citet{raffel2019exploring} and \citet{goodwin2020towards}.
    



To assess the value of using multi-dataset sampling, in our experiments we will compare the original proportional mixing and temperature scaling by \citet{raffel2019exploring} with our approach. 

\section{Experimental Settings}

General figures for all the datasets are reported in Table \ref{table:data_split}, while more detailed textual statistics are available in Appendix \ref{sec:appendix_a}. 
More details about the training can be found in Appendix \ref{sec:appendix_b}. 

\subsection{Datasets}

\paragraph{SMM4H} This dataset was introduced for the Shared Tasks on AE in the Workshop on Social Media Mining for Health Applications (SMM4H) \citep{weissenbacher2018overview}. The dataset is composed of Twitter posts, typically short, informal texts with non-standard ortography, and it contains annotations for both detection (i.e., Task 1, classification) and extraction (i.e., Task 2, NER) of AEs. The number of samples differs from the original dataset as many tweets vanished, due to deletion or access restriction in the platform. Splits are stratified, to maintain an equal ratio of positive and negative examples (see Table \ref{table:data_split}).

\paragraph{CADEC} CADEC contains 1,250 medical forum posts annotated with patient-reported AEs. In this dataset, texts are long and informal, often deviating from English syntax and punctuation rules. Forum posts may contain more than one AE. For our goals, we adopted the training, validation, and test splits proposed by \citet{dai2020effective} (see Table \ref{table:data_split}).

\paragraph{ADE corpus v2} This dataset \cite{gurulingappa2012development} contains case reports extracted from MEDLINE and it was used for multi-task training, as it contains annotations for all tasks in Table \ref{table:tasks}, i.e. drugs, dosage, AE detection and extraction. Splits are stratified, to maintain an equal ratio of positive and negative examples (see Table \ref{table:data_split}).

\paragraph{WEB-RADR} This dataset is a manually curated benchmark based on tweets. We used it exclusively to test the performance of the multi-task models, as it was originally introduced only for testing purposes \cite{dietrich2020adverse} (see Table \ref{table:data_split}).

\begin{table}[!ht] 
\renewcommand\thetable{2}
\resizebox{\columnwidth}{!}{%
\begin{small}
\begin{tabular}{llll}
\hline
Dataset                                                                                     & Total & Positive   & Negative \\ \hline
\begin{tabular}[c]{@{}l@{}}\textbf{SMM4H Task 1}\\ \textbf{(AE Detection)}\end{tabular}&   15,482     &   1,339   &   14,143     \\
Train (80\%)                                                                                & 12,386 & 1,071 & 11,315 \\
Validation (10\%)                                                                           & 1,548   & 134  & 1,414   \\
Test (10\%)                                                                                 & 1,548   & 134  & 1,414   \\ \hline
\begin{tabular}[c]{@{}l@{}}\textbf{SMM4H Task 2}\\ \textbf{(AE Det., AE \& Drug Extr.)}\end{tabular}     &    2,276    &  1300    &    976    \\
Train (60\%)                                                                                & 1,365   & 780  & 585    \\
Validation (20\%)                                                                           & 455    & 260  & 195    \\
Test (20\%)                                                                                 & 456    & 260  & 196    \\ \hline
\begin{tabular}[c]{@{}l@{}}\textbf{CADEC}\\ \textbf{(AE Det., AE \& Drug Extr.)}\end{tabular}      &    1,250    &  1,105    &      145  \\
Train (70\%)                                                                                & 875    & 779  & 96     \\
Validation (15\%)                                                                           & 187    & 163  & 24     \\
Test (15\%)                                                                                 & 188    & 163  & 25     \\ \hline
\begin{tabular}[c]{@{}l@{}}\textbf{ADE Corpus v2}\\ \textbf{(AE Detection)}\end{tabular}    & 23,516       &  6,821  &   16,695         \\
Train (60\%)                                                                                    & 14,109  & 4,091 & 10,018  \\
Validation (20\%)                                                                               & 4,703   & 1,365 & 3,338   \\
Test (20\%)                                                                                     & 4,704   & 1,365 & 3,339   \\ \hline
\begin{tabular}[c]{@{}l@{}}\textbf{ADE Corpus v2}\\ \textbf{(AE Extraction)}\end{tabular}          &     6,821   &   6,821   &   0     \\
Train (60\%)                                                                                    & 4,091   & 4,091 & 0      \\
Validation (20\%)                                                                               & 1,365   & 1,365 & 0      \\
Test (20\%)                                                                                     & 1,365   & 1,365 & 0      \\ \hline
\begin{tabular}[c]{@{}l@{}}\textbf{ADE Corpus v2}\\ \textbf{(Drug Extraction)}\end{tabular}        &    7,100    &  7,100    &      0  \\
Train (60\%)                                                                                    & 4,260   & 4,260 & 0      \\
Validation (20\%)                                                                               & 1,420   & 1,420 & 0      \\
Test (20\%)                                                                                     & 1,420   & 1,420 & 0      \\ \hline
\begin{tabular}[c]{@{}l@{}}\textbf{ADE Corpus v2}\\ \textbf{(Drug Dosage Extraction)}\end{tabular} &    279    &   0   &    0    \\
Train (60\%)                                                                                    & 167    & 0    & 0       \\
Validation (20\%)                                                                               & 56     & 0    & 0       \\
Test (20\%)                                                                                     & 56     & 0    & 0       \\ \hline
\begin{tabular}[c]{@{}l@{}}\textbf{WEB-RADR}\\ \textbf{(AE Detection \& Extraction)}\end{tabular}      &       &    &      \\
Test                                                                                        & 57,481   & 1,056 & 56,425    \\  \hline
\begin{tabular}[c]{@{}l@{}}\textbf{SMM4H-French}\\ \textbf{(AE Detection)}\end{tabular}      &       &    &     \\
Test                                                                                        & 1,941   & 31  & 1,910    \\  \hline
\end{tabular}%
\end{small}
}
\caption{Dataset Statistics and Splits.}
\label{table:data_split}
\end{table}

\paragraph{SMM4H-French} The SMM4H French Dataset contains a total of 1,941 samples out of which 31 samples belong to AE (positive) class and 1,910 samples have the label Non-AE (negative class). This dataset is only used for testing the zero-shot transfer (see Table \ref{table:data_split}).

\subsection{Settings}

\paragraph{AE Detection} We train and test T5 and the baselines (see \ref{baseline:ae_detect}) on the SMM4H Task 1 dataset. We then assess the robustness of T5 and the best performing baseline on the test sets of CADEC, ADE Corpus v2 and WEB-RADR.

\paragraph{AE Extraction} We train and test T5 and the baselines (see \ref{baseline:ae_extract}) on the SMM4H Task 2 dataset. We then assess the robustness of T5 and the best performing baseline by testing them (trained on either SMM4H Task 2 or CADEC) on the test sets of SMM4H Task 2, CADEC, ADE Corpus v2 and WEB-RADR.

\paragraph{Multi-Task Learning} We train T5-Base on all the training sets for all tasks, using proportional mixing or temperature scaling both with the original multi-task (see \ref{baseline:ae_multi}) and with our proposed multi-task and multi-dataset approach, and we evaluate the resulting models on the available test sets.

\paragraph{Language Transfer} We train T5 and the Multilingual BERT (see \ref{baseline:ae_crossling}) 
on the SMM4H Task 1 English dataset, and then we test it in a zero-shot learning setting on the SMM4H-French dataset.


\subsection{Baselines}
\label{sec:baselines}

\subsubsection{AE Detection}
\label{baseline:ae_detect}
Our baselines are five pre-trained BERT variants with a classification head fine-tuned for AE detection. A weighted cross-entropy loss function is used for all of them to adjust for class imbalance.\\
\textbf{BioBERT} \cite{lee2020biobert} was built upon the original BERT and further pre-trained on PubMed abstracts. We used BioBERT v1.1, which was reported to perform better in biomedical tasks. \\
\textbf{BioClinicalBERT} \cite{alsentzer2019publicly} was pre-trained on MIMIC III dataset containing Electronic Health Records (EHR) of ICU patients. \\
\textbf{SciBERT} \cite{beltagy2019scibert} was pre-trained on 1.14 million papers, randomly selected from semantic scholar, with an 18-82 ratio between computer science and biomedical papers.\\  
\textbf{PubMedBERT} \cite{gu2020domain} was pre-trained from scratch on PubMed abstracts, without building upon the vocabulary of the original BERT. \\
\textbf{SpanBERT} \cite{joshi2020spanbert} adopts a different pre-training objective from BERT. This model is trained by masking full contiguous spans instead of single words or subwords, which allows it to encode span-level information. 

\subsubsection{AE Extraction Task Baselines}
\label{baseline:ae_extract}
For the \textsc{AE extraction} task, we use the four models described in \citet{portelli-etal-2021-bert}, 
namely BERT, BERT+CRF, SpanBERT, and SpanBERT+CRF. The authors reported state-of-the-art performance with the SpanBERT models on SMM4H, and their implementation is publicly available at \url{https://github.com/ailabUdineGit/ADE}.  

\subsubsection{Multi-Task Learning}
\label{baseline:ae_multi}
For Multi-Task Learning, we use as baseline the T5 model fine-tuned with the original training strategies by \citet{raffel2019exploring}, which balance across tasks (TB, task balancing) but do not account for multi-dataset learning (DB, dataset balancing). We refer to them as T5\textsubscript{TB-PM} for proportional mixing and T5\textsubscript{TB-TS} for temperature scaling. We refer to our approach, which accounts also for the multi-dataset learning, as T5\textsubscript{TDB-PM} for proportional mixing and T5\textsubscript{TDB-TS} for temperature scaling.

\subsubsection{Language Transfer}
\label{baseline:ae_crossling}
As a baseline for Language Transfer, we use Multilingual BERT (the uncased version), which was pre-trained on monolingual corpora in 102 languages \citep{devlin2018bert}. The model was fine-tuned by adding a classification head on the top to perform AE Detection in a zero-shot  setting. 

\subsection{Metrics} We adopt the same metrics of the SMM4H competition. \footnote{\url{https://competitions.codalab.org/competitions/20798}} For the AE Detection (i.e., the \textit{assert ade} prefix) we use precision, recall, and F1-score for the positive (AE) class.  For the AE Extraction (i.e., the \textit{ner ade}, \textit{ner drug}, \textit{ner dosage} prefixes) we use both Strict and Partial Match F1-Score \citep{weissenbacher2019overview,klein2020overview}. The same AE Detection and AE Extraction metrics have also been used in the Multi-Task setting and in the Language Transfer settings.

\section{Results and Analysis}

\subsection{AE Detection}

Table \ref{table:assert_ade_results} summarizes precision, recall and F1 score obtained by T5-Small, T5-Base and the baselines on the SMM4H Task 1 test set. 

\begin{table}[!ht]
\centering
\begin{tabular}{lccc} 
\hline
\textbf{Model}   & \textbf{Precision}  & \textbf{Recall}  & \textbf{F1}   \\ 
\hline
BioBERT   & 55.5                & 63.1             & 59.0                \\
BioClinicalBERT  & 68.3                & 59.7             & 63.7                \\
SciBERT        & 68.8                & 55.9             & 61.7               \\
PubMedBERT    & 59.7                & 61.9             & 60.8                \\
SpanBERT       & 55.0                & 73.1             & 62.8                \\
T5-Small       & 58.1                & 65.0             & 61.3                    \\
\textbf{T5-Base}        & \textbf{68.8}                & \textbf{73.7}             & \textbf{71.1}                \\
\hline
\end{tabular}
\caption{Precision, Recall and F1-Score for the positive AE class in the SMM4H Task 1 test set}
\label{table:assert_ade_results}
\end{table}


T5-Small obtains competitive performance, lagging slightly behind the performance of some BERT variants, while T5-Base outperforms all the other approaches, with a 12.7\% relative F1-score improvement over the best baseline, BioClinicalBERT (the improvement for the McNemar test is significant at $p < 0.001$).
It should also be noticed that the two versions of T5, together with SpanBERT, improve over the Recall of the other BERT variants. The result seems to comply with the report by \citet{portelli-etal-2021-bert,beatrice2021span}, who found that models relying on span-based objectives had increased recall in the task, probably because they are better at identifying longer AE spans that would otherwise go undetected.

\begin{table}[!ht]
\resizebox{\columnwidth}{!}{%
\begin{small}
\begin{tabular}{lcccc}
\hline
\begin{tabular}[c]{@{}c@{}}\textbf{Model}\textbackslash{}\\\textbf{Test set}\end{tabular} & \begin{tabular}[c]{@{}c@{}}\textbf{SMM4H} \\ \textbf{Task 2}\end{tabular} & \textbf{CADEC} & \begin{tabular}[c]{@{}c@{}}\textbf{ADE}\\ \textbf{Corpus}\\ \textbf{v2}\end{tabular} & \begin{tabular}[c]{@{}c@{}}\textbf{WEB-}\\ \textbf{RADR}\end{tabular} \\ \hline
BioClinical BERT                                                                & 82.5                                                    & 90.1  & 28.6                                                       & 32.3                                                 \\ 
T5-Base                                                                      & \textbf{88.0}                                                    & \textbf{93.7}  & \textbf{31.7}                                                       & \textbf{35.8}                                                     \\ \hline
\end{tabular}%
\end{small}
}
\caption{F1-Score for T5-Base and BioClinicalBERT trained on SMM4H Task 1 and tested on all datasets.}
\label{table:ttg_assert_ade}
\end{table}

Table \ref{table:ttg_assert_ade} provides the results for the model generalization evaluation that we run for T5 and the best baseline. In this evaluation, we train the systems on SMM4H Task 1 and test on the other datasets (i.e.  SMM4H Task 2, CADEC, ADE Corpus V2 and WEB-RADR), which differ from the training set in terms of linguistic features, text structures, text lengths and even annotation schemes. Both models obtain high performance on SMM4H Task 2 and CADEC, despite their textual differences. The large linguistic difference of the ADE Corpus v2 (i.e., MEDLINE case reports) explains instead the drastic drop in performance for both systems in this dataset, in which T5-Base still performs better than the baseline. WEB-RADR also proves to be a challenging benchmark for its extreme class imbalance, but  our system still achieves an F1-score around 0.36 for the positive class, while BioClinicalBERT is performs than the T5-Base.

\begin{figure}[!bp]
\centering
    \includegraphics[scale=0.26]{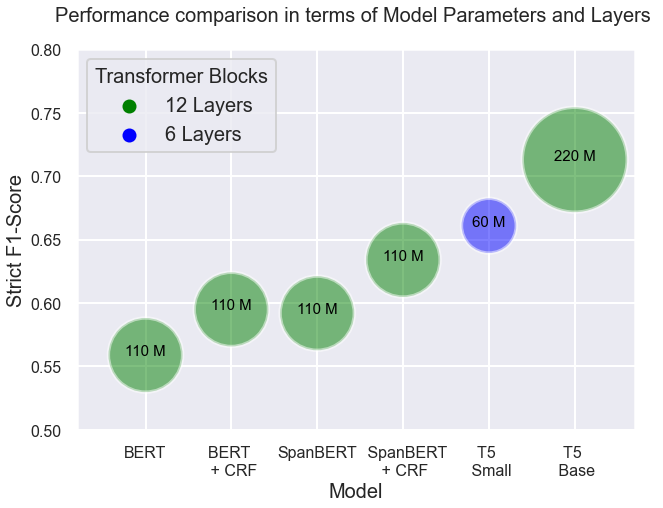}
    \caption{Performance in the AE Extraction task, with the number of layers and parameters for each system.}
   \label{fig:param_f1}
\end{figure}

\begin{table*}[!ht]
\centering
\begin{tabular}{lcccc}
\hline
\multicolumn{1}{c}{} & \multicolumn{2}{c}{\textbf{SMM4H Task 2}}                   & \multicolumn{2}{c}{\textbf{CADEC}} \\ \hline
Architecture         & Partial F1 & \multicolumn{1}{c|}{Strict F1} & Partial F1   & Strict F1  \\ \hline
BERT                 & 66.1       & \multicolumn{1}{c|}{55.9}      & 77.7         & 65.2       \\
BERT+CRF             & 68.1       & \multicolumn{1}{c|}{59.5}      & 77.2         & 64.4       \\
SpanBERT             & 66.7       & \multicolumn{1}{c|}{59.2}      & 79.2         & 67.2       \\
SpanBERT + CRF       & 70.1       & \multicolumn{1}{c|}{63.4}      & \textbf{79.4}         & 67.6       \\ \hline
T5-Small             & 70.7       & \multicolumn{1}{c|}{66.1}      & 75.6         & 65.7           \\
T5-Base              & \textbf{75.1}       & \multicolumn{1}{c|}{\textbf{71.3}}      & 79.1         & \textbf{69.8}            \\ \hline
\citet{dai2020effective}  & -       & \multicolumn{1}{c|}{-}      & -        & 69.0            \\ \hline
\end{tabular}
\caption{Partial and Strict F1 score for the AE Extraction task on SMM4H Task 2 and CADEC. For CADEC, we also report the current SOTA model by \citet{dai2020effective}.
}
\label{table:ae_extraction_res}
\end{table*}

\subsubsection{Qualitative Analysis on SMM4H}
 In order to better understand the model performance, we picked some samples from the SMM4H Task 1 test dataset to compare between captured and non-captured AE and analyze the reason behind the miss-classification. In few cases, the model has problems identifying non-standardized acronyms, for example the input ``really bad RLS from <drug name>'', is classified as non-AE by the model compared to its original label as an AE. The model is not able to understand the meaning behind RLS, which denotes Restless Leg Syndrome in this scenario. We observed that if the RLS is changed to nightmares, headache or restless leg syndrome, the model recognizes the input as an AE. The model is able to capture most of the AE unusual references such as ``<drug name> burns like thousand suns'', ``<drug name> was a joke'', ``<drug name> tastes like battery acid''. Yet we found some cases in which the model failed. For example, the inputs ``stomach feels like a cement mixer after taking <drug name>'' was classified as non-AE. In this case, ``cement mixer'' is used in a figurative way to refer to the fact that the stomach is not well or it is churning. Once we replace this figurative image with a term such as churning, the model correctly classifies the sample as AE. 
\subsection{AE Extraction}

Table \ref{table:ae_extraction_res} summarizes the results for the AE Extraction task for T5 and the baselines trained on SMM4H Task 2, including the scores for a recent SOTA system on CADEC \citep{dai2020effective}. It can be seen that both T5 models outperform all the baselines on the SMM4H data, while on the longer and more structured CADEC texts the SpanBERT architectures are more competitive for the partial F1-score. On the other hand, our best model still retains a better performance for the Strict F1 metric, suggesting that it is more accurate in detecting the boundaries of the AE span. T5-Base also outperforms the system by \citet{dai2020effective}.


\begin{table}[!ht]
\resizebox{\columnwidth}{!}{%
\begin{small}
\begin{tabular}{lcccc}
\hline
\multicolumn{1}{c}{Model}                                & \begin{tabular}[c]{@{}c@{}}SMM4H \\ Task 2\end{tabular} & CADEC       & \begin{tabular}[c]{@{}c@{}}ADE\\ Corpus\\ v2\end{tabular} & \begin{tabular}[c]{@{}c@{}}WEB-\\ RADR\end{tabular} \\ \hline
\multicolumn{5}{l}{\textit{Trained on SMM4H Task 2}}                                                                                                                                                                                               \\
\begin{tabular}[c]{@{}l@{}}SpanBERT\\ + CRF\end{tabular} & 70.1 (63.4)                                             & 15.7 (2.8)  &  24.6 (15.1)                                                     & 18.9 (7.3)                                          \\\hline
T5-Base                                                  & \textbf{75.1 (71.3)}                                             & \textbf{24.4 (20.5)} & \textbf{38.9 (29.5)}                                               & \textbf{36.2 (13.9)}                                                    \\ \hline \hline
\multicolumn{5}{l}{\textit{Trained on CADEC}}                                                                                                                                                                                                      \\
\begin{tabular}[c]{@{}l@{}}SpanBERT\\ + CRF\end{tabular} & 35.4 (28.6)                                                         & 79.4 (67.6)  &  31.2 (24.8)                                                          & 20.1 (7.9)                                                    \\\hline
T5-Base                                                  & \textbf{57.9 (51.6)}                                             & 79.1 (69.8) & \textbf{50.3 (43.7)}                                               & \textbf{30.4 (18.8)}                                                    \\ \hline
\end{tabular}%
\end{small}
}
\caption{Partial (strict) F1-scores for T5-Base and SpanBERT+CRF trained on SMM4H Task 2 and CADEC and evaluated on all datasets.}
\label{table:ttg_ae_extract}
\end{table}

In order to further evaluate the system generalization capability, we test on all the AE Extraction datasets both T5-Base and SpanBERT+CRF (best baseline), after training them on SMM4H Task 2 and on CADEC. In Table \ref{table:ttg_ae_extract}, it can be seen that T5-Base has better generalization than the baseline on all datasets, with F1-scores that are 10 points higher or more. Training on CADEC generalizes better (with the only exception of the partial metric for WEB-RADR), while systems trained on the SMM4H perform poorly on the other benchmarks.

Fig. \ref{fig:param_f1} compares the baselines and the T5 performance in AE extraction, in terms of number of layers/parameters. The plot suggests that the model parameters and the number of layers are not the factors for the T5 models performance gain, e.g. T5-Small has almost half the number of parameters (60 million) and half of the layers (6 layers) of BERT and its variants and it still performs better.

\begin{table*}[!ht]
\resizebox{\textwidth}{!}{%
\begin{tabular}{c|ccccc}
\hline
\textbf{Text Statistics}                  & \textbf{BERT} & \textbf{BERT+CRF} & \textbf{SpanBERT} & \textbf{SpanBERT+CRF} & \textbf{T5-Base} \\ \hline
Dale Chall Readability$^+$ & 8.34          & 8.15              & 8.20              & 8.32                  & \textbf{9.44}             \\
Automated Readability$^+$  & 8.42          & 8.37              & 8.35              & 8.51                  & \textbf{10.37}            \\
Flesch Reading Ease$^-$    & 62.73         & 63.57             & 62.60             & 62.92                 & \textbf{53.18}   \\ \hline        
\end{tabular}%
}
\caption{Text Statistics metric to evalute the quality of span generated by models trained on SMM4H Task 2 dataset ( $^+$ represents higher score is better and $^-$ means lower score is better)}
\label{table:text_stats}
\end{table*}

\begin{table*}[!ht]
\begin{small}
\resizebox{\textwidth}{!}{%
\begin{tabular}{l|ccccccc}
\hline
\multicolumn{1}{c|}{\textbf{Task}}                & \textbf{Model/Dataset} & \textbf{\begin{tabular}[c]{@{}c@{}}\underline{SMM4H} \\ \underline{Task 1}\end{tabular}} & \textbf{\begin{tabular}[c]{@{}c@{}}\underline{SMM4H} \\ \underline{Task 2}\end{tabular}} & {\ul \textbf{CADEC}} & \textbf{\begin{tabular}[c]{@{}c@{}}\underline{ADE} \\ \underline{Corpus v2}\end{tabular}} & \textbf{\begin{tabular}[c]{@{}c@{}}\underline{WEB-} \\ \underline{RADR}\end{tabular}} & \multicolumn{1}{l}{{\ul \textbf{Avg. Score}}} \\ \specialrule{1.5pt}{1pt}{1pt}
\multicolumn{1}{c|}{\multirow{4}{*}{assert  ade}} & T5\textsubscript{TB -PM}            & 55.2                                                             & \textbf{91.5}                                                    & 92.7                 & 91.7                                                              & 31.9                                                & 72.6                                \\
\multicolumn{1}{c|}{}                             & T5\textsubscript{TDB - PM}   & \textbf{67.9}                                                    & 88.5                                                             & \textbf{98.7}        & 91.7                                                             & \textbf{37.4}                                                          & \textbf{76.8}                                         \\ \cline{2-8} 
\multicolumn{1}{c|}{}                             & T5\textsubscript{TB -TS}            & 65.3                                                             & 83.5                                                             & 91.1                 & 90.9                                                              & 36.1                                                          & 73.3                                          \\
\multicolumn{1}{c|}{}                             & T5\textsubscript{TDB - TS}   & \textbf{69.4}                                                    & \textbf{89.4}                                                    & \textbf{98.7}        & \textbf{91.5}                                                     & \textbf{37.3}                                                 & \textbf{77.2}                                \\ \specialrule{1.5pt}{1pt}{1pt}
\multirow{4}{*}{ner ade}                          & T5\textsubscript{TB -PM}            & -                                                                & 75.7 (71.8)                                                             & 46.5 (39.9)                & 58.4 (53.4)                                                              & 38.6 (15.1)                                                         & 54.8 (45.0)                                        \\
                                                  & T5\textsubscript{TDB - PM}   & -                                                                & 75.7 (71.8)                                                             & \textbf{74.4 (64.0)}        & \textbf{59.7 (55.9)}                                                     & \textbf{38.7 (15.8)}                                                 & \textbf{62.1 (51.8)}                                 \\ \cline{2-8} 
                                                  & T5\textsubscript{TB -TS}            & -                                                                & 75.3 (70.2)                                                             & 45.2 (38.4)                 & 59.7 (56.1)                                                              & 38.9 (15.6)                                                         &        54.7 (45.0)                                          \\
                                                  & T5\textsubscript{TDB - TS}   & -                                                                & \textbf{75.7 (71.1)}                                                    & \textbf{75.3 (66.0)}        & \textbf{60.3 (56.7)}                                                     & \textbf{39.1 (15.8)}                                                 & \textbf{62.6 (52.4)}                                 \\ \specialrule{1.5pt}{1pt}{1pt}
\multirow{4}{*}{ner drug}                         & T5\textsubscript{TB -PM}            & -                                                                & \textbf{92.3 (92.3)}                                                    & 88.7 (88.7)                 & 79.4 (79.0)                                                              & -                                                             & 86.8 (86.6)                                        \\
                                                  & T5\textsubscript{TDB - PM}   & -                                                                & 90.3 (90.3)                                                             & \textbf{92.4 (91.8)}        & \textbf{82.2 (82.0)}                                                     & -                                                             & \textbf{88.3 (88.0)}                                 \\ \cline{2-8} 
                                                  & T5\textsubscript{TB -TS}            & -                                                                & 88.2 (88.1)                                                             & 88.4 (88.1)                 & 80.2 (79.8)                                                              & -                                                             & 85.6 (85.3)                                          \\
                                                  & T5\textsubscript{TDB - TS}   & -                                                                & \textbf{91.8 (91.8)}                                                    & \textbf{94.1 (93.4)}        & \textbf{83.1 (82.8)}                                                     & -                                                             & \textbf{89.6 (89.3)}                                 \\ \specialrule{1.5pt}{1pt}{1pt}
\multirow{4}{*}{ner dosage}                       & T5\textsubscript{TB -PM}            & -                                                                & -                                                                & -                    & 73.2 (67.8)                                                              & -                                                             & 73.2 (67.8)                                         \\
                                                  & T5\textsubscript{TDB - PM}   & -                                                                & -                                                                & -                    & \textbf{78.5 (71.4)}                                                     & -                                                             & \textbf{78.5 (71.4)}                                 \\ \cline{2-8} 
                                                  & T5\textsubscript{TB -TS}            & -                                                                & -                                                                & -                    & 76.7 (71.4)                                                              & -                                                             & 76.7 (71.4)                                          \\
                                                  & T5\textsubscript{TDB - TS}   & -                                                                & -                                                                & -                    & \textbf{78.5 (71.4)}                                                              & -                                                             & \textbf{78.5 (71.4)}                                          \\ \specialrule{1.5pt}{1pt}{1pt}
\end{tabular}%

}
\end{small}
\caption{F1-scores for the multi-task setting. Task Balancing (TB) is compared to our Task and Dataset Balancing (TDB) approach, with PM $=$ Proportional Mixing and TS $=$ Temperature Scaling. F1 of the positive class is reported for AE Detection (the \textit{assert ade} row), while \textit{partial (strict)} F1 is reported for the Extraction tasks.}
\label{table:multi_task}
\end{table*}


\subsubsection{Analysis of Extracted Spans on SMM4H}
We employ some commonly used text statistics to assess the spans extracted by the T5 model. Table \ref{table:text_stats} compares three text statistics metrics for the model trained on the SMM4H Task 2 dataset. The higher scores obtained by T5 in the Dale Chall Readability \cite{chall1995readability} and Automated Readability index \cite{smith1967automated} suggest this model is able to generate a higher percentage of AE spans with rare terms. The lower Flesch Reading score \cite{kincaid1975derivation}, instead, indicates that the model generates spans that are more readable.

\subsection{Multi-Task Learning}
\label{sec:multitask_results}

Table \ref{table:multi_task} includes the scores on all the test sets 
for the multi-task T5 models, trained either with the original or with our proposed strategy (see \ref{sec:training_strategies}).

In AE Detection, our T5\textsubscript{TDB} approach always outperforms the original T5\textsubscript{TB} by a large margin (5.8\% relative improvement for PM and 5.3\% for TS), except for the Proportional Mixing case in SMM4H Task 2.
Margins are smaller in ADE Corpus V2 and WEB-RADR. Looking at the comparison between TS and PM for T5, the former is better in the SMM4H subsets and comparable in all the others, globally obtaining a higher average score.

Our training approach improves both partial and strict F1-scores on the AE Extraction task, where the models are tested on all datasets except for SMM4H Task 1, which does not have AE Extraction annotations (13.3\% relative improvement for PM and 14.4\% for TS). In all datasets, our training strategies obtain equal or superior performance for both partial and strict F1 scores, with large gains on CADEC and more marginal gains on SMM4H Task 2, ADE Corpus v2 and WEB-RADR. TS is again preferable to PM, obtaining a higher average score.
The results for the Drug and Dosage tasks are similar: in Drug Extraction, SMM4H Task 2 confirms to be more challenging for T5\textsubscript{TDB-PM} (T5\textsubscript{TB-PM} outperforms it by 2 points), while T5\textsubscript{TDB-TS} outperforms its counterpart. In all the other settings, the Task and Dataset Balancing approaches score higher than Task Balancing-only ones.

Overall, our approaches consistently achieve gains in the multi-task setting, independently from task type (i.e. Detection or Extraction) and annotation scheme. 
TS proves to be superior to PM in all tasks, even though it may lag slightly behind PM in some datasets. 


\subsection{Cross-lingual Transfer}

As a final evaluation, we tested the ability of T5-Base and Multilingual BERT to generalize the AE Detection task to a new language, i.e. French. Notice that the SMM4H French data proved to be challenging, due to the extreme class imbalance \citep{klein2020overview}. It can be seen in Table \ref{table:zero_shot} 
that T5-Base obtains higher F1-score, specifically thanks to a higher precision. Multilingual BERT, instead, shows higher recall. Overall, the T5-Base performance in zero-shot learning is encouraging, and further improvements are likely to come with few shot learning or with more targeted strategies for multilingual training.

\begin{table}[!ht]
\centering
\begin{small}
\begin{tabular}{cccc}
\hline
Architecture      & \multicolumn{3}{c}{\underline{Zero-Shot}} \\
                  & Precision   & Recall  & F1    \\ \hline
Multilingual BERT & 10.2        & \textbf{32.2}    & 15.5  \\
T5-Base           & \textbf{17.9}        & 22.6    & \textbf{20.0}  \\ \hline
\end{tabular}%
\caption{Metrics for Multilingual BERT and T5-Base on zero-shot learning on SMM4H-French.} 
\label{table:zero_shot}
\end{small}
\end{table}





\section{Conclusions}
In order to address several typical challenges of the healthcare domain (small, imbalanced and highly variable datasets, cross-lingual data), we proposed to treat AE Detection and AE/Drug/Dosage Extraction tasks as sequence-to-sequence problems, adapting the T5 architecture and improving over all the baselines in both the Detection and the Extraction tasks. To maximize the benefit of multi-task and multi-dataset learning, we introduced a new training strategy that extends \citet{raffel2019exploring}, showing that our approach accounts for multiple and diverse datasets and leads to consistent improvements over the original T5 proposal. Finally, the model also shows some language transfer abilities in the zero shot setting, leaving the door open for future experiments to extend our training framework towards multilinguality \citep{xue2020mt5}.  

\section*{Acknowledgments}
We would like to thank the reviewers and the chairs for their insightful reviews and suggestions.



\bibliography{anthology,acl2020, cit}
\bibliographystyle{acl_natbib}

\appendix

\newpage
\section{Dataset Textual Statistics}
\label{sec:appendix_a}
Table \ref{table:data_diff} presents textual statistics to show the difference in type of datasets with respect to their input sequence length, target (extraction) span sequence length and other parameters. It can observed that the input sequence length is relatively short for the SMM4H and for WEB-RADR datasets, while CADEC and ADE corpus datasets tend to include longer texts. The Flesch reading ease score \citep{flesch1949art} indicates the readability of the sentence, with lower values representing that the text is difficult to understand for the average reader. The ADE corpus datasets have the lowest Flesch reading score, as the text is adopted from MEDLINE and contains more medical terms, while Twitter data (SMM4H, WEB-RADR) and the health forum (CADEC) datasets contain a lower amount of scientific terminology and are typically made of shorter texts, with a lower degree of syntactic complexity.  

\label{sec:data_diff}
\begin{table*}[!bp]
\resizebox{\textwidth}{!}{%
\begin{tabular}{lcccccccc}
\hline
\textbf{Dataset}                                                                   & \textbf{\begin{tabular}[c]{@{}c@{}}Avg. Seq\\ Length\end{tabular}} & \textbf{\begin{tabular}[c]{@{}c@{}}Avg. \\ Span Length\\ (AE, Drug or \\ Dosage)\end{tabular}} & \textbf{\begin{tabular}[c]{@{}c@{}}Avg. \\ Stopwords\\ in span\end{tabular}} & \textbf{\begin{tabular}[c]{@{}c@{}}Avg. Freq. \\ of AE per \\ sample\end{tabular}} & \textbf{\begin{tabular}[c]{@{}c@{}}Unique AE\\ words\end{tabular}} & \textbf{\begin{tabular}[c]{@{}c@{}}\% of AE\\ Samples\end{tabular}} & \textbf{\begin{tabular}[c]{@{}c@{}}Unique Drug\\ Mentions\end{tabular}} & \textbf{\begin{tabular}[c]{@{}c@{}}Flesch Reading\\ Ease Score\end{tabular}} \\ \hline
\begin{tabular}[c]{@{}l@{}}SMM4H Task 1\\ (AE Detection)\end{tabular}              & 98.9                                                               & -                                                                                              & -                                                                            & -                                                                               & -                                                                  & 8.6                                                                 & -                                                                       & 64.7                                                                         \\ \hline
\begin{tabular}[c]{@{}l@{}}SMM4H Task 2\\ (AE Det., AE \& Drug Extr.)\end{tabular} & 108.8                                                              & 9.1                                                                                            & 0.2                                                                          & 1                                                                               & 1108                                                               & 57.1                                                                & 69                                                                      & 62.1                                                                         \\ \hline
\begin{tabular}[c]{@{}l@{}}CADEC\\ (AE Det., AE \& Drug Extr.)\end{tabular}        & 459.4                                                              & 16.1                                                                                           & 2.4                                                                          & 6                                                                               & 2303                                                               & 89.0                                                                & 320                                                                     & 69.1                                                                         \\ \hline
\begin{tabular}[c]{@{}l@{}}ADE Corpus v2\\ (AE Detection)\end{tabular}             & 132.5                                                              & -                                                                                              & -                                                                            & -                                                                               & -                                                                  & 28.9                                                                & -                                                                       & 23.2                                                                         \\ \hline
\begin{tabular}[c]{@{}l@{}}ADE Corpus v2\\ (AE Extraction)\end{tabular}            & 152.1                                                              & 18.5                                                                                           & 0.1                                                                          & 1                                                                               & 2662                                                               & 100                                                                 & -                                                                       & 13.6                                                                         \\ \hline
\begin{tabular}[c]{@{}l@{}}ADE Corpus v2\\ (Drug Extraction)\end{tabular}          & 152.3                                                              & 10.8                                                                                           & 0                                                                            & -                                                                               & -                                                                  & 100                                                                 & 1251                                                                    & 14.3                                                                         \\ \hline
\begin{tabular}[c]{@{}l@{}}ADE Corpus v2\\ (Drug Dosage Extraction)\end{tabular}   & 163.4                                                              & 8.5                                                                                            & 0                                                                            & -                                                                               & -                                                                  & 100                                                                 & -                                                                       & 23.6                                                                         \\ \hline
\begin{tabular}[c]{@{}l@{}}WEB-RADR\\ (AE Detection \& Extraction)\end{tabular}    & 106.3                                                              & 16.5                                                                                           & 1.1                                                                          & 2                                                                               & 2037                                                               & 1.8                                                                 & -                                                                       & 61.3                                                                         \\ \hline
\begin{tabular}[c]{@{}l@{}}SMM4H French\\ (AE Detection)\end{tabular}              & 142.4                                                              & -                                                                                              & -                                                                            & -                                                                               & -                                                                  & 1.6                                                                 & -                                                                       & -                                                                            \\ \hline
\end{tabular}%
}
\caption{Comparison of the AE datasets according to different textual statistics.}
\label{table:data_diff}
\end{table*}

\section{Training Details}
\label{sec:appendix_b}
All the experiments have been performed on the top of Hugging-face’s Python package ~\cite{wolf2019huggingface}. \footnote{\href{https://github.com/huggingface/transformers}{https://github.com/huggingface/transformers}} The code for the models implemented in the paper is available at \url{https://github.com/shivamraval98/MultiTask-T5_AE}
\subsection{AE Detection}
The baseline BERT models for AE detection were trained on one NVIDIA Tesla V100 16 GB GPU and it takes the model approximately 30 minutes to execute for all epochs. The hyperparameters used for baseline models are detailed in Table \ref{table:ae_detection_bert}.
\begin{table}[!h]
\resizebox{\columnwidth}{!}{%
\begin{tabular}{l c c c }
\textbf{Model}           & \textbf{Epoch} & \textbf{Batch Size} & \textbf{Warm-up Steps} \\ \hline
\textbf{BioBERT}         & 3              & 32                  & 400                   \\
\textbf{BioClinicalBERT} & 5              & 40                  & 500                   \\
\textbf{SciBERT}         & 5              & 40                  & 400                   \\
\textbf{PubMedBERT}      & 5              & 40                  & 300                   \\
\textbf{SpanBERT}        & 3              & 40                  & 400                  
\end{tabular}%
}
\caption{Hyperparameters for AE Detection baselines. The learning rate and weight decay was kept constant with values $5e-05$ and $0.01$ respectively}
\label{table:ae_detection_bert}
\end{table}

The T5 models were trained using a cluster of four NVIDIA Tesla V100 16 GB GPU, with 80 batch size per GPU and 10 epochs for T5-Small, and 16 batch size per GPU and 7 epochs for T5-Base. The learning rate for the both the t5 models was set to $1e-04$. The input and the generated sequence length were set to 130 and 20, respectively, with exponential length penalty set to 2 for the generated sequence. For the rest of the hyperparameters, we used the default values in the library. 

The T5-Small model approximately takes 3-5 minutes per epoch while T5-Base executes for 7-10 minutes per epoch in the aforementioned cluster environment setting. 

\subsection{AE Extraction}
The hyperparameters for the baseline models (BERT, BERT+CRF, SpanBERT and SpanBERT+CRF) of AE extraction were set as described in \citet{portelli-etal-2021-bert}. The hyperparameter setting for the T5-Small and T5-Base for both SMM4H Task 2 and CADEC dataset is presented in Table \ref{table:ae_ext} and the default values were utilized for the rest of the hyperparameters.

\begin{table}[!ht]
\resizebox{\columnwidth}{!}{%
\begin{tabular}{lcccccc}
\textbf{Model} & \textbf{ISL} & \textbf{OSL} & \textbf{BS} & \textbf{EP} & \textbf{LR} & \textbf{Time} \\ \hline
\multicolumn{7}{l}{\textit{SMM4H Task 2 AE Extraction}}                                                \\
T5-Small       & 130          & 20           & 80          & 10          & 1e-4        & 5             \\
T5-Base        & 130          & 20           & 64          & 7           & 1e-4        & 7            \\ \hline
\multicolumn{7}{l}{\textit{CADEC AE Extraction}}                                                       \\
T5-Small       & 512          & 150          & 64          & 25          & 1e-3        & 10            \\
T5-Base        & 512          & 150          & 32          & 20          & 1e-3        & 20          \\ \hline
\end{tabular}%
}
\caption{Hyperparameters for T5-Small and T5-Base when trained on SMM4H and CADEC AE Extraction Task (ISL = Input Sequence Length, OSL = Output Sequence Length, BS = Batch Size (over all GPU's), EP = Epoch, LR = Learning Rate, Time = Training Time in mins per epoch). }
\label{table:ae_ext}
\end{table}

\subsection{Multi-Task Training}
The Multi-Task Training was performed on T5-Base by combining all the training sets and experimenting for the originally proposed Task Balancing (TB) approach, and for our proposed task plus multi-dataset balancing (TDB) strategy for proportional mixing (PM) and temperature scaling (TS). The same hyperparameters were utilized for all settings with batch size $8$, learning rate $1e-04$, input sequence length $512$ and output sequence length $150$. Temperature value was kept to be 2 for the temperature scaling method. For every multi-task setting, it took the model approximately 60 minutes to train for one epoch in the 4 GPU cluster computing environment setting. 

\subsection{Cross-Lingual Transfer}
Multilingual BERT was trained using the four GPU cluster setting with batch size $256$ over all GPU's for 7 epochs. The learning rate was set as $5e-05$ with $0$ warmup steps and $0.01$ weight decay. The T5-Base model trained on English SMM4H Task 1 AE Detection dataset was utilized to perform zero-shot on SMM4H French Dataset.

\end{document}